\documentclass[runningheads]{llncs}
\usepackage[T1]{fontenc}
\usepackage{graphicx}
\usepackage{amssymb}
\usepackage{amsmath}
\usepackage{graphicx}
\usepackage{array}
\usepackage{paralist}
\usepackage{booktabs}
\usepackage{comment}
\usepackage{multirow}
\usepackage{hyperref}

\usepackage[tableposition=top]{caption}

\usepackage[table,dvipsnames]{xcolor}

\usepackage{todonotes}
\usepackage[autolanguage]{numprint}
\usepackage{capt-of}

\usepackage{arydshln}

\begin{document}
\title{Probing Pretrained Language Models with Hierarchy Properties}
\author{Jesús Lovón-Melgarejo $^{\spadesuit}$, Jose G. Moreno $^{\spadesuit}$, Romaric Besançon $^{\diamondsuit}$\\ { Olivier Ferret $^{\diamondsuit}$, Lynda Tamine $^{\spadesuit}$ } \\
${}^{\spadesuit}$Université Paul Sabatier, IRIT, Toulouse, France\\
${}^{\diamondsuit}$Université Paris-Saclay, CEA, List, Palaiseau, France\\
{\small \tt \{jesus.lovon, jose.moreno, tamine\}@irit.fr}\\
{\small \tt \{romaric.besancon, olivier.ferret\}@cea.fr}
}

\institute{}

\maketitle              
\begin{abstract}

Since Pretrained Language Models (PLMs) are the cornerstone of the most recent Information Retrieval (IR) models, the way they encode semantic knowledge is particularly important. However, little attention has been given to studying the PLMs' capability to capture hierarchical semantic knowledge.
Traditionally, evaluating such knowledge encoded in PLMs relies on their performance on a task-dependent evaluation approach based on proxy tasks, such as hypernymy detection. Unfortunately, this approach potentially ignores other implicit and complex taxonomic relations. In this work, we propose a task-agnostic evaluation method able to evaluate to what extent PLMs can capture complex taxonomy relations, such as ancestors and siblings. The evaluation is based on intrinsic properties that capture the hierarchical nature of taxonomies. Our experimental evaluation shows that the lexico-semantic knowledge implicitly encoded in PLMs does not always capture hierarchical relations. We further demonstrate that the proposed properties can be injected into PLMs to improve their understanding of hierarchy. 
Through evaluations on taxonomy reconstruction, hypernym discovery  and reading comprehension tasks, we show that the knowledge about hierarchy is moderately but not systematically transferable across tasks.


\end{abstract}
\vspace{-0.9cm}
\section{Introduction}
The hierarchical representation of concepts is a fundamental aspect of human cognition and essential in performing numerous  
  Information Retrieval (IR) (e.g., web search \cite{sieg2004}) and   Natural Language Processing (NLP) tasks (e.g., hypernym discovery \cite{camacho-collados-etal-2018-semeval}).

Therefore, works in this research direction have explored incorporating hierarchical knowledge extracted from knowledge graphs and taxonomies to refine models \cite{hu_2015,liu2019kbert,nickel_2017,chen-etal-2021-CTP}. 

In particular, prior studies on static word embeddings showed that designing dedicated space representations to encode the hierarchy benefits the performance of various downstream tasks \cite{nickel_2017,vendrov_2016a}, giving rise to the importance of encoding concept hierarchy in these representations.\\

Most of previous studies have adopted task-dependent evaluations to assess the extent to which models capture hierarchical linguistic knowledge. These evaluations are based on the premise that the model's performance on downstream tasks sheds light on whether the model captures hierarchy. Specifically, such evaluations target tasks that require applying hierarchical knowledge to some degree, such as taxonomy reconstruction and hypernymy detection \cite{nickel_2017,camacho-collados_2017}. 

However, existing task-dependent evaluations have two main limitations. First, they heavily rely on detecting a single relation type (hypernymy) and overlook the implicit taxonomy structure by ignoring other related relations, such as ancestors and siblings \cite{mao-etal-2018-end}. As a result, none of the state-of-the-art (SOTA) evaluation methodologies are able to reveal this implicit yet essential hierarchical information \cite{alsuhaibani2019joint}. 

Second, particularly in the context of Pretrained Language Models (PLMs), task-dependent evaluations might conflate the model's understanding of a given target task and the model's understanding of hierarchy per se. 

Hence, there is a need for more comprehensive evaluation methods and datasets that consider complex hierarchical relations and are able to reveal  how well models capture those relations and apply them to downstream tasks.

The present work addresses the two aforementioned limitations. We propose a task-agnostic methodology  to evaluate language models' understanding of hierarchical knowledge considering implicit and more complex taxonomic relations beyond the direct hypernymy (parent, ancestors, and siblings). 
In particular,  our evaluation focuses on PLMs, such as BERT \cite{devlin_2019}, considering their ability to encode lexical knowledge, as well as their outstanding performance on different tasks \cite{Qiu-2020}. 
To the best of our knowledge, no task-agnostic methodology exists so far that enables us to reveal to what extent PLMs encode hierarchy relations intrinsically.
Following previous work on information retrieval \cite{camara2020diagnosing} and lexical semantics \cite{talmor_2020,vulic_2020}, we use a \textit{probe}-based methodology \cite{pimentel-etal-2020-information} to evaluate whether SOTA models capture task-agnostic linguistic knowledge.

The article is structured as follows. We first introduce a set of intrinsic properties of a hierarchy based on edge-distance observations in a taxonomy. Then, we design probes consisting of triplets of entities, named  \textit{ternaries}, that encode these properties and are used to evaluate and teach PLMs a notion of hierarchy. \\

We aim to answer three research questions. Firstly, \textbf{(RQ1)} \textit{To which extent do PLM
representations encode hierarchy w.r.t. hierarchy properties?} To address this question, we analyze PLM representations with our task-agnostic evaluation based on taxonomic relations between concepts.  
Secondly, \textbf{(RQ2)} \textit{Does injecting hierarchy properties into PLMs using a task-agnostic methodology impact their representations?} where we fine-tuned hierarchy-enhanced PLMs  with probes built upon the hierarchy properties and evaluated them. 
Lastly, \textbf{(RQ3)}\textit{Can hierarchy-enhanced PLM representations be transferred to downstream tasks?} where we evaluated the performance impact of hierarchy-enhanced PLMs on Hypernym Discovery, Taxonomy Reconstruction, and Reading Comprehension. \\
Our main findings regarding these questions are: 1)  the evaluated PLMs struggle to capture hierarchical relations, such as siblings and ancestor representations;
 2) PLMs can enhance their representations by learning the hierarchy properties; these enhanced models improve their performance on our evaluation; 3) There appears to be a gap in understanding between hierarchy knowledge and task, making it difficult to achieve a clear trend of performance increase of enhanced PLMs across downstream tasks.


\vspace{-1.0em}
\section{Related Work and Background}

\subsection{Hierarchical Representations: Learning and Evaluation}
Learning knowledge representations with an underlying hierarchical structure is an active research topic \cite{rossi_2021}. Approaches include learning embedding representations from symbolic knowledge sources (e.g., knowledge graphs (KGs)) such as TransE \cite{bordes_2013} and TransR \cite{ji-etal-2015-knowledge} or constructing dedicated vector spaces based on hyperbolic spaces \cite{nickel_2017} and order embeddings \cite{vendrov_2016a}.   

Recently, with the emergence of PLMs, different works explored infusing factual knowledge from KGs. 
The infusion is typically done by injecting triplets embeddings to capture entity and relation features \cite{liu2019kbert,zhang_2019}. However, current approaches are limited to the triplet representation and risk only to capture partial graph structure information \cite{yang2022survey}. In our work, we use a \textit{ternary} of entities to represent three entities implicitly linked between them with a hierarchy-based relation, aiming to build more hierarchy-aware representations.

To assess the quality of model representations, a frequent evaluation task is Taxonomy Reconstruction \cite{nickel_2017,camacho-collados_2017}.
Multiple frameworks extended this evaluation with different features of a taxonomy. For instance, evaluations are based on the granularity and generalization paths \cite{gupta_2016} or structural features such as cycles, connected components, and intermediate nodes \cite{bordea_2015,bordea_2016}.
Similarly, recent work studied PLMs' adaptation to Taxonomy Reconstruction \cite{jain_2022} and Hypernymy Detection tasks via sequence classification \cite{chen-etal-2021-CTP}. Additionally, Chen et al. \cite{chen2020probing}  explored the syntactic and semantic hierarchies at the term-level of BERT within the hyperbolic space tailored to the sentiment classification task. However, our methodology differs from this approach. We conceive and analyze the notion of hierarchy between concepts, translating it into a set of properties that provide an intuitive and robust basis for evaluation. These properties, based on established semantic similarity metrics from SOTA \cite{budanitsky_2006,resnik_1995}, impose constraints on the relations between concepts. Finally, while non-Euclidean spaces for hierarchy exploration have garnered recent attention \cite{du2022hakg,song2022hyperbolic}, they fall beyond our current scope.

\vspace{-0.5em}
\subsection{Probes on PLMs}
\vspace{-0.5em}

We consider the probes as challenging datasets, not conceived to provide new knowledge but to assess what a model already knows about a task, typically under a light fine-tuning or zero-shot setup \cite{richardson_2020}. In the context of PLMs, different efforts are proposed to unveil what kind of knowledge these models encode using these probes. Recent work has explored probes for information retrieval \cite{camara2020diagnosing}, natural language inference \cite{clark_2020}, symbolic reasoning \cite{talmor_2020}, and lexical semantics \cite{vulic_2020}. 
Other approaches \cite{richardson_2020,talmor_2020} developed evaluation methodologies, including probe evaluations before and after fine-tuning these models on target data samples. Under this approach, it is observable how adaptable a model is to the learning probe format. In our work, we apply probes in a zero-shot setup to study the default model PLM representations.

\section{A Task-Agnostic Methodology: Design and Evaluation}

This section presents our task-agnostic methodology for probing hierarchical knowledge of PLMs. 

First, we define a set of intrinsic properties for a hierarchy (§3.1). Then, we describe how to design probes upon these properties and how to fine-tune PLMs to identify these properties by learning these probes (§3.2). 
\vspace{-0.5em}
\subsection{Intrinsic Hierarchy Properties}
\vspace{-0.5em}
We characterize a hierarchical structure with a set of intrinsic properties that mirror the distribution of concepts within a given taxonomy.
From previous work on lexical resources (e.g., taxonomies \cite{zhong_2002,budanitsky_2006}), we represent concepts as nodes and consider edge-based approaches, such as the shortest path between two nodes, to define a notion of similarity between concepts.
Concretely, we define a set of \textit{relations} and \textit{properties} from edge-based observations. Then, we evaluate these properties empirically considering semantic distance methods, inspired by \cite{budanitsky_2006}\footnote{We consider distance as the complement metric for similarity.}. 

Notably, for a fixed node $n$, we define four basic relations --\textit{parent}, \textit{ancestor}, \textit{sibling}, and \textit{far relative}-- in a taxonomy in the following way. A \textit{parent} node $p$ is a direct hypernym at a one-edge distance from $n$, while an \textit{ancestor} node is an indirect hypernym at a two-edge distance. A \textit{sibling} node shares a parent with a two-edge distance, and a \textit{far relative} shares the ancestor but not the parent.

We use these relations and the corresponding edge-based distances to define hierarchy properties. Table~\ref{tab:list_properties} presents the six hierarchy properties considered in this work based on the four defined relations for all possible combinations. In order to verify these properties, we formulate them as inequalities that examine the distance between two distinct relations for a fixed node, as shown in Table~\ref{tab:properties}. 

Our motivation for considering three nodes (\textit{ternary}) in these evaluations comes from recent research on pattern-based relation extraction \cite{hovy-etal-2009-toward,liu2023seeking}, where the inclusion of a third ``anchor'' node has proven useful in capturing various relation types, including hypernymy and co-hyponymy.

To facilitate the reading, we adopt a naming convention for the properties that highlights the two used relations, i.e. a \textbf{R}elation-\textbf{R}elation format. The relation identifier, denoted by \textbf{R}, can take one of four possible values: \textbf{P}arent, \textbf{A}ncestor, \textbf{S}ibling, and \textbf{F}ar relative. Furthermore, we congregate the initial properties into three \textit{groups}, P-*, A-*, and S-*, based on the left relation in the inequality.  These \textit{groups} serve as an aggregated representation of the properties for each type of relation.

\begin{table}[!htb]
    \scriptsize
    \centering
    \caption{\small The six hierarchy properties with their names and textual descriptions.}
    
    \begin{tabular}{cp{10.5cm}}
    \toprule
     \small \textbf{Property} & \small \textbf{Description}\\
     \midrule
      \textit{\textbf{P-A}}    &  \emph{``A  node is closer to its parent than its ancestor.''} The similarity between a pair of concepts is proportional to their path length under an edge-based approach \cite{resnik_1995}. 
     This property also follows the intuition of over-generalization edges at the evaluation of granularity  \cite{gupta_2016}, limited to only hypernym/hyponym relations  \\ \midrule
     \textit{\textbf{P-S}}& \emph{``The distance between `siblings' should be longer than between `father' and `son'.''}. Similarly, it is a straightforward application of edge-counting, also found in other approaches such as scaling the taxonomy \cite{zhong_2002}. \\ \midrule
     \textit{\textbf{P-F}} & \emph{``The parent is the closest element to a node.''} It generalizes P-S by comparing to further relations in a taxonomy.  
   If a model does not satisfy this property, it struggles to differentiate the hypernym relation from others.\\ \midrule 
\textit{\textbf{A-S}} & \emph{``A  node is closer to its ancestor than to its sibling.''} 
    We did not find an expected behavior about this property in the literature. We empirically choose the path-level evaluation proposed by \cite{gupta_2016}, favoring the correct edges in the hypernym path rather than adding incorrect elements that could cause a cascade of generalization errors.\\ \midrule
    \textit{\textbf{A-F}} & \emph{``The ancestor is the closest term for a node, except for the father.''} This property is a generalization of the ancestor relationship evaluated in further relationships in a taxonomy. \\ \midrule
    \textit{\textbf{S-F}} & \emph{``The sibling is the closest element for a node, except for the father and the ancestor.''} Based on edge-counting approaches, we should find a sibling node closer to other relations beyond the ancestor in a hierarchy.\\
     \bottomrule
    \end{tabular}
    \label{tab:list_properties}
\end{table}

%

\begin{table}[!htb]
    \scriptsize
    \caption{ \small The hierarchy properties along with their distance-based definitions, and the three participant (colored) nodes in the taxonomy. Three \textit{groups} are identified based on the left relation in the inequality: \colorbox{NavyBlue!25}{P-*} (regrouping P-A, P-S, and P-F), \colorbox{YellowGreen!40}{A-*} (regrouping A-S and A-F), and \colorbox{YellowOrange!25}{S-*} (with only S-F). }
    \centering
    \resizebox{\textwidth}{!}{
    \begin{tabular}{ccc|ccc}
    \toprule
       \textbf{Property} & \multicolumn{2}{c}{\textbf{Definition}}  & \textbf{Property} & \multicolumn{2}{c}{\textbf{Definition}} \\ \midrule
         \cellcolor{NavyBlue!25}\textbf{P-A} & $ dist(n,p) < dist(n,a) $&  \begin{minipage}{.13\textwidth}
      \includegraphics[width=0.9\linewidth]{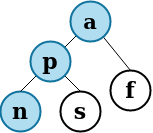}
    \end{minipage}  &
           \cellcolor{YellowGreen!40}\textbf{A-S} & $ dist(n,a) < dist(n,s) $&\begin{minipage}{.13\textwidth}
      \includegraphics[width=0.9\linewidth]{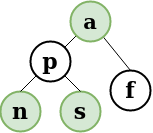}
    \end{minipage}\\ 
         \cellcolor{NavyBlue!25}\textbf{P-S}& $ dist(n,p) < dist(n,s) $ &\begin{minipage}{.13\textwidth}
      \includegraphics[width=0.9\linewidth]{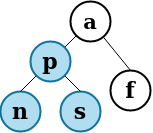}
    \end{minipage}   &
    \cellcolor{YellowGreen!40}\textbf{A-F} & $ dist(n,a) < dist(n,f) $ &\begin{minipage}{.13\textwidth}
      \includegraphics[width=0.9\linewidth]{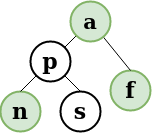}
    \end{minipage}\\
         
    \cellcolor{NavyBlue!25}\textbf{P-F}& $ dist(n,p) < dist(n,f) $ &\begin{minipage}{.13\textwidth}
      \includegraphics[width=0.9\linewidth]{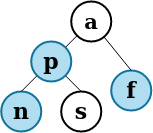}
    \end{minipage}&
          \cellcolor{YellowOrange!25}\textbf{S-F} & $ dist(n,s) < dist(n,f) $  &\begin{minipage}{.13\textwidth}
      \includegraphics[width=0.9\linewidth]{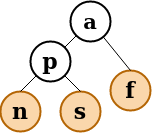}
    \end{minipage}\\
         \bottomrule
    \end{tabular}
}
    \label{tab:properties}
\end{table}

\subsection{Probing Hierarchical Representations} \label{sec:3.2}

\paragraph{\textbf{Designing probes.}}
For this purpose, we align the properties into a single form.
Given a taxonomy $T=(V,E)$, we define a \textit{ternary} as $t=(x_n,x_l,x_r)$ encoding a hierarchy property \textbf{R$_l$-R$_r$} where $x_n$ represents a fixed node, $x_l$ the node of the \textbf{l}eft \textbf{R}elation (\textbf{R$_l$}), and $x_r$ the node of the \textbf{r}ight \textbf{R}elation (\textbf{R$_r$}). Note that each of the tuples $(x_n,x_l)$ and $(x_n,x_r)$ corresponds to one of the defined relations (parent \textbf{P}, ancestor \textbf{A}, sibling \textbf{S} or far relative \textbf{F}) and the taxonomic distances (edge-based) must satisfy the associated property $dist(x_n,x_l)<dist(x_n,x_r)$. 

Each node within the ternary tuple is converted into textual representations through predefined phrases (prompts) (§4.3). These concept representations are used to compute a model representation within a given language model. Consequently, for a given model, we obtain a representation in the form $(\hat{x_n}, \hat{x_l}, \hat{x_r})$ for each ternary $(x_n,x_l,x_r)$. Finally, we use a distance method to evaluate the inequality $d(\hat{x_n},\hat{x_l}) < d(\hat{x_n}, \hat{x_r})$.

\paragraph{\textbf{Teaching hierarchy properties to a PLM.}}
To teach the hierarchy properties to PLMs, we leverage the concept representations derived from the probes, $(\hat{x_n}, \hat{x_l}, \hat{x_r})$, and train a model to satisfy the inequality $d(\hat{x_n},\hat{x_l}) < d(\hat{x_n}, \hat{x_r})$.  We assume that a model's performance on these probes indicates how well its representations align to a hierarchy-like distribution. We employ the Sentence Transformer framework \cite{reimers_2019} with a triplet network architecture and a pooling operation to the output of the PLM to generate the embedding of our concepts. This approach achieved promising results on lexical knowledge evaluations \cite{vulic_2021}.

Given the inner contrastive nature of our probes in a triplet form, we adopt a contrastive loss, specifically the Triplet loss \cite{dong_2018a}. This loss fine-tunes the network to minimize the distance between related inputs, $(\hat{x_n}$, $\hat{x_l})$, while maximizing the distance for unrelated inputs, $(\hat{x_n}, \hat{x_r})$, by minimizing the following loss:

\begin{equation}
    \mathcal{L}(x_n,x_l,x_r) = - max(\lVert \hat{x_n}-\hat{x_l} \rVert - \lVert \hat{x_n}-\hat{x_r} \rVert + \alpha, 0)
    \label{eq:loss}
\end{equation}

\section{Experimental Setup}

\subsection{Datasets and Metrics}
We used the Bansal dataset to sample our probes, consisting of medium-sized taxonomies generated from WordNet subtrees \cite{bansal_2014}. This dataset comprises subtrees of height 3 (i.e., the longest path from the root to the leaf is 4 nodes) containing between 10 and 50 terms. Specifically, we used the version extended with WordNet definitions \cite{chen-etal-2021-CTP}.

We generated ternaries for each property, splitted into train/dev partitions to fine-tune PLMs with the hierarchy properties. We generated approximately 20,000/4,000/14,000 for each property's train/dev/test split, except for \textit{P-A} with 6,300/1,400/1,400 ternaries. As mentioned before, we do not consider the property \textit{A-S} for training due to its absence in the literature. For evaluation, we use the accuracy metric defined as the number of correct predictions where the ternary inequality was satisfied, divided by the total number of ternaries in the test set.

\vspace{-1.0em}
\subsection{Baselines and Models}
As baselines, we use three groups of models: 

\textit{Group 1.} Comprises random and non-contextualized models as a lower bound performance of the PLMs following previous work \cite{vulic_2021,talmor_2020}: a)  \textbf{Random:} we generated symmetrical random distances for all node pairs and report the average of ten random runs; b) \textbf{FastText:} to compare static word embeddings in a fair setup, i.e. avoiding the out-of-vocabulary problem, we used the character-based version of \textbf{FastText} embeddings \cite{bojanowski_2017} trained on Wikipedia (\textbf{FT-wiki}). 

 \textit{Group 2.} Comprises a total of five PLMs available in the HuggingFace library: 
 a) \textbf{BERT} (\textit{bert-base-cased}); b) \textbf{BERT-L} (\textit{bert-large-cased}); c) \textbf{RoB-L} (\textit{roberta-large}); d) \textbf{S-RoB} (\textit{all-distilroberta-v1}); and e) \textbf{S-MPNet} (\textit{all-mpnet-base-v2}).
We notice that the first three models are cross-encoders and the last two models (\textbf{S-RoB} and \textbf{S-MPNet}) are bi-encoders. The bi-encoders are trained using a dual-encoder network and oriented towards semantic textual similarity tasks on multiple datasets. These characteristics give some advantages in the final results of our probes compared to classical PLMs.   However, we evaluated these models to obtain insights on the most robust ones.
 
 \textit{Group 3.} Comprises two SOTA knowledge enhanced PLMs: \textbf{ERNIE} \cite{Sun_Wang_Li_Feng_Tian_Wu_Wang_2020}, a BERT-based trained on multiple tasks to capture lexical, syntactic and semantic aspects of information, and \textbf{CTP} \cite{chen-etal-2021-CTP}, a RoBERTa large model trained to perform hypernym prediction.

\vspace{-1.0em}
\subsection{Concept and Ternary Representations}

Each concept in a ternary is given a textual representation that consists of  concept's name  and its definition to enrich context information. This representation is a list of tokens in the form \textit{`[CLS] [concept name]  is defined as [definition] [SEP]'}. We investigate different methods for generating the representation of this textual form from a PLM, including vectorial-based (\textbf{cls} and \textbf{avg}) and prompt-based methods. 
For vectorial-based methods, we consider the last layer output. The  \textbf{cls} method uses the special token \textit{[CLS]} and the \textbf{avg} method computes the average over all the subtokens. To compute the distance between these representations, we consider cosinus (\textbf{cos}) and euclidean (\textbf{L2}) distances. 
For the prompt-based method, we use the LMScorer method \cite{jain-starsem-22}, which scores sentences for factual accuracy. LMScorer computes a pseudo-likelihood score for each token in a sequence by iteratively masking it, considering past and future tokens: $ LMScorer_{PLM}(\mathbf{W}) = \exp \left(\sum_{i=1}^{|\mathbf{W}|} log P_{PLM}(w_i | \mathbf{W}_{\setminus i})\right)$. 
We experiment with different templates for converting ternaries into textual representations, reporting the best-scoring template\footnote{Template: ``A is a B. C is a B. [definition a][definition b] [definition c].'', where A, B, and C are nodes.}. For static word embeddings, we use only the \textbf{avg} method, as the \textit{[CLS]} token is not available.

\vspace{-0.5em}
\section{Results and Analysis}

In this section, we report the results  to answer our research questions.
To simplify our analysis, we use the \textit{groups} presented in Table \ref{tab:properties} and report the average values.
We only report the best configuration for each model due to space limitations, but we ensure we obtain the best configuration for each model.\footnote{Data, models and training details will be publicly available at : \url{https://github.com/xxxx}}

\begin{table}[t]
\centering
\scriptsize
\setlength{\tabcolsep}{6pt}
\caption{\small Accuracy for Property \textit{P-A} with different representation methods. LMScorer is not computed for bi-encoder models since they are not adapted for this task \cite{wang-etal-2021-tsdae-using}.}

\begin{tabular}{ccccccc}

\toprule
\textbf{Model} & \textbf{Distance} &  \multirow{2}{*}{BERT} &  \multirow{2}{*}{BERT-L} &  \multirow{2}{*}{RoB-L} &   \multirow{2}{*}{S-RoB} &  \multirow{2}{*}{S-MPNet} \\
\textbf{Representation} & \textbf{Method}\\
\midrule
\multirow{2}{*}{cls} & cos   &  72.6 &    68.2 &     58.5 &    83.1 &   83.2 \\
 &L2 &  72.4 &    68.7 &     58.5 &     79.1 &   83.1 \\ \hline 
\multirow{2}{*}{avg} & cos &  \textbf{76.0} &    \textbf{78.9} &         \textbf{75.0} &    \textbf{85.2} &   \textbf{85.6}\\
&L2 &  75.7 &    78.0 &     74.8 &    80.6 &   \textbf{85.6} \\ \hline 
LMScorer&-& 50.2 & 52.6 &  54.6 & - & -\\

\bottomrule
\vspace{-1.5em}
\end{tabular}

\label{tab:methods}
\end{table}


\vspace{-1.0em}
\subsection{Evaluation of Hierarchy Properties in PLMs}

First, we carry out preliminary experiments to explore the best performing settings in terms of model representation and distance methods. 
Table \ref{tab:methods} shows our results based only on the representative property \textit{P-A} using various methods on all PLMs. Our findings indicate that the vector-based representations (75.0-85.6) outperform all the explored prompts. We also observe that the \textbf{avg} representation is always superior to \textbf{cls}. Moreover, the \textbf{avg} representation with \textbf{cos} distance frequently obtains slightly better scores (76.0) than \textbf{L2} (75.7). Under this evidence, we adopted the \textbf{avg}  representation with \textbf{cos} distance for our remaining evaluations.

We now answer our first research question \textbf{RQ1:} \textit{To which extent do PLM
representations encode hierarchy w.r.t. hierarchy properties?}  
Our analysis focuses on the model's overall performance in our evaluation (named \textit{All}, as the average score from all properties) and provides insights into the property \textit{groups} (\textit{P-*}, \textit{S-*}, and \textit{F-*}) captured by   \textit{Group 1} and \textit{Group 2} models.

Considering the model's overall performance (\textit{All}), we first test the quality of our probes from our baselines in \textit{Group 1}. Table \ref{tab:rq1} shows that \textbf{FastText (FT-wiki}) embeddings perform better ($60.3$) than the \textbf{Random} approach ($50.2$). Aligned with previous work \cite{fu_2014} claiming that static embeddings encode hypernymy-like relations to some degree, we argue that our probes help to capture these hierarchical relations. 
Similarly, all \textit{Group 2} models obtained \textit{All} scores higher than \textbf{Random}. In particular, the \textbf{S-RoB} model obtained the highest score ($70.3$), closely followed by \textbf{S-MPNet} ($70.0$). Moreover, only these models  outperformed the \textbf{FT-wiki} static embeddings.  From \textit{Group 3} models, we can surprisingly see that \textbf{ERNIE} outperforms \textbf{CPT}, while the latter is a RoBERTa large model specifically trained on hypernym classification task with a classification layer on top of the PLM.
By comparing these models against their vanilla version, we can interestingly observe that while  \textbf{ERNIE} showed constant improvement in all properties, with $+2$ score points on \textit{All} ($56.2$) w.r.t. \textbf{BERT}, our probes suggest that \textbf{CTP}, trained on a task-dependent evaluation ($37.7$), degrades its initial representations w.r.t \textbf{RoB-L} ($53.3$). 
Moreover, \textbf{CTP} is the only PLM-based model with lower performance than \textbf{Random}. Overall, these results provide insights on the conflation problem faced with task-dependent evaluation regarding task understanding and hierarchy understanding of PLM's.

\begin{table}[tb]
 \caption{\small Accuracy on each hierarchy property on the test dataset with method \textit{avg(cos)}. Best results in \textbf{bold} for each probe per group. * shows 1 sample t-test $>0.05$.}
\scriptsize

\centering
\begin{tabular}{lp{0.08\textwidth}p{0.075\textwidth}p{0.075\textwidth}p{0.075\textwidth}p{0.075\textwidth}p{0.075\textwidth}p{0.07\textwidth}p{0.07\textwidth}p{0.07\textwidth}p{0.07\textwidth}}
\toprule
    Model &   \textbf{P-A} &   \textbf{P-S} &   \textbf{P-F} &     \textbf{A-S} &   \textbf{A-F} &    \textbf{S-F} &   \textbf{P-*}& \textbf{A-*}&\textbf{S-*}&\textbf{All}\\
\midrule
& \multicolumn{10}{c}{\scriptsize\textbf{\textit{Group 1: Baselines}}}\\
Random & 50.5 & 49.8 & 50.1 & 50.1 & 50.4 & 50.1 & 50.1 & 50.3 & 50.1 & 50.2 \\
  FT-wiki & 79.7 & 49.9 & 81.6 & 24.1 & 48.5 & 78.2 & \textbf{70.4} & 36.3 & \textbf{78.2} & \textbf{60.3} \\ \midrule
   &   \multicolumn{10}{c}{\scriptsize\textbf{\textit{Group 2: Vanilla PLMs}}}\\
     BERT &  75.8 &  37.9 &  71.8 &  22.3 &  43.0 &  74.5 &  61.8 &  32.6 &  74.5 &  54.2 \\
  BERT-L &  78.8 &  42.0 &  77.6 &  24.6 &  47.6 &  76.1 &  66.1 &  36.1 &  76.1 &  57.8 \\
RoB-L &  73.2 &  37.2 &  71.1 &  24.6 &  40.1 &  73.8 &  60.5 &  32.4 &  73.8 &  53.3\\
 S-RoB &  85.2 &  68.6 &  90.4 &  33.7 &  64.8 &  78.8 &  \textbf{81.4} & \textbf{49.2} &  78.8 &  \textbf{70.3}\\
    S-MPNet &  85.6 &  70.0 &  88.3 &  29.6 &  65.9 &  80.6 &  81.3 &  47.8 &  \textbf{80.6} &  70.0 \\ \midrule
        &  \multicolumn{10}{c}{\scriptsize\textit{\textbf{Group 3: KG-Enhanced PLMs }}}\\
    ERNIE &  77.0 &  42.7 &  74.2 &  23.1 &  44.9 &  75.6 &  \textbf{64.6} &  34.0 &  \textbf{75.6} &  \textbf{56.2}\\
      CTP &  59.9 &  22.4 &  36.2 &  23.9 &  22.8 &  61.0 &  39.5 &  23.4 &  61.0 &  37.7 \\

  \midrule
          &  \multicolumn{10}{c}{\scriptsize\textit{\textbf{Group 4: Hierarchy-Enhanced PLMs}}}\\
  BERT$_{hp}$ & 79.9  & 56.2  & 82.6  & 29.2  & 58.4 & 75.2 & 72.9 & 43.8 & 75.2 & 64.0* \\
  BERT-L$_{hp}$ & 85.7 & 69.1 & 91.0 & 31.6 & 67.6 & 79.3 & 81.9 & 49.6 & 79.3 & 70.3*\\ 
  RoB-L$_{hp}$ & 82.0 & 64.3 & 86.0  & 35.1 & 61.9 & 76.1 & 77.4 & 48.5 & 76.1 & 67.4* \\
  S-RoB$_{hp}$ & 87.7 & 74.9 & 93.9 & 34.5 & 73.0 & 81.6 & 85.5 & 53.8 & \textbf{81.6} & \textbf{73.6}* \\
  S-MPNet$_{hp}$ & 84.0 & 80.0 & 92.9 & 35.5 & 72.3 & 80.4 &\textbf{85.6} &\textbf{53.9} & 80.4 & 73.3* \\

\bottomrule
\end{tabular}
\vspace{-1.2em}
\label{tab:rq1}
\end{table}

Now, we analyze the results at \textit{groups} level. 
Considering the semantic similarity approach in pre-training, we expect that \textit{sibling} relations (\textit{S-*}), akin to analogies, will exhibit better representations than other hierarchical relations.

Across all models, we can observe that the \textit{siblings} relation (\textit{S-*}), followed by the \textit{parent} relation (\textit{P-*}), obtained higher performances than the \textit{ancestor} relation (\textit{A-*}). 
Our results capture the preference of the semantic similarity of these models,
particularly when compared to other \textit{far} relations, because of the high values in \textit{S-F} ($73.8-80.6$).
Similarly, most \textit{ancestor} representations are further than \textit{parents} ($73.2-85.6$ in \textit{P-A}). Based on property \textit{P-S} ($37.2-42.0$), PLMs such as \textbf{BERT}, \textbf{BERT-L}, and \textbf{RoB-L} tend to represent \textit{siblings} closer ($73.8-76.1$) than the \textit{parent} ($60.5-66.1$).
For the \textit{ancestor} relation, higher scores on property \textit{A-F} ($40.1-65.9$) than on \textit{A-S} ($22.3-33.7$) imply that \textit{far relatives} are easier than \textit{siblings}. However, these scores are generally lower than those obtained with other properties. Besides, when comparing the \textit{P-F} and \textit{A-F} columns, we observe higher scores with the \textit{parent}, one-edge distance, than the \textit{ancestor}, two edges away (similarly, for the scores between \textit{P-S} and \textit{A-S}). These trends indicate that \textit{ancestor} representations are not as clearly defined as \textit{parents}. \\
  
\vspace{-1.0em}
Overall, our evaluation reveals that the analyzed PLMs struggle to capture some hierarchical relationships such as \textit{sibling} and  \textit{ancestor}. Besides, the results confirm that the performance of PLM specifically trained on a hierarchy-aware task (e.g., \textbf{CTP})  is not a salient signal of PLM's understanding  of hierarchy.

\vspace{-1.0em}
 \subsection{Enhancing PLMs with hierarchy properties}
 \vspace{-0.5em}
In the following, we answer \textbf{RQ2:} \textit{Does injecting hierarchy properties into PLMs using a task-agnostic methodology impact their representations?} Our underlying objective is to investigate the feasibility of our task-agnostic evaluation of PLMs w.r.t to hierarchy through the set of defined properties (§3.1). To this end, we fine-tuned (§3.2) and evaluated the models on the probes following the best evaluation setup for PLMs (discussed in §5.1). These fine-tuned models are referenced as hierarchy-enhanced PLMs (denoted \textit{PLM$_{hp}$}) belonging to \textit{Group 4} in Table \ref{tab:rq1}, where we report the average accuracy after fine-tuning each model with 5 different random seeds and optimal hyper-parameters. Similar to RQ1, we consider the \textit{All} for comparison, and we take as reference the corresponding vanilla model for our analysis. 

Our results showed improvement in all models for the \textit{All} column w.r.t the original PLM.
For instance, the models \textbf{BERT-L$_{hp}$}, \textbf{S-RoB$_{hp}$}, and \textbf{S-MPNet$_{hp}$} are improved by $12.5$, $3.3$, and $3.3$ points, respectively in comparison to their vanilla counterparts. 
In particular, the property \textit{A-S}, not included in the training, shows a constant improvement for all models. These findings align with our hypothesis that \textit{A-S} complements other property criteria without compromising initial performance. In contrast, the trained property \textit{S-F} shows a more modest improvement ($+1.8$ on average), implying that while our fine-tuning approach yields promising results, it still struggles to differentiate \textit{sibling} and \textit{far relation} representations effectively.

\begin{figure}[tb]
    \caption{ \small Avg. distance in the property evaluation for properties \textit{P-S} and \textit{A-S} (left). One example for each property, showing the distances of different concepts for a fixed node with models S-RoB and S-RoB$_{hp}$. We indicate the $\Delta$ distance between concepts (right).}
    \centering
    \includegraphics[width=\textwidth]{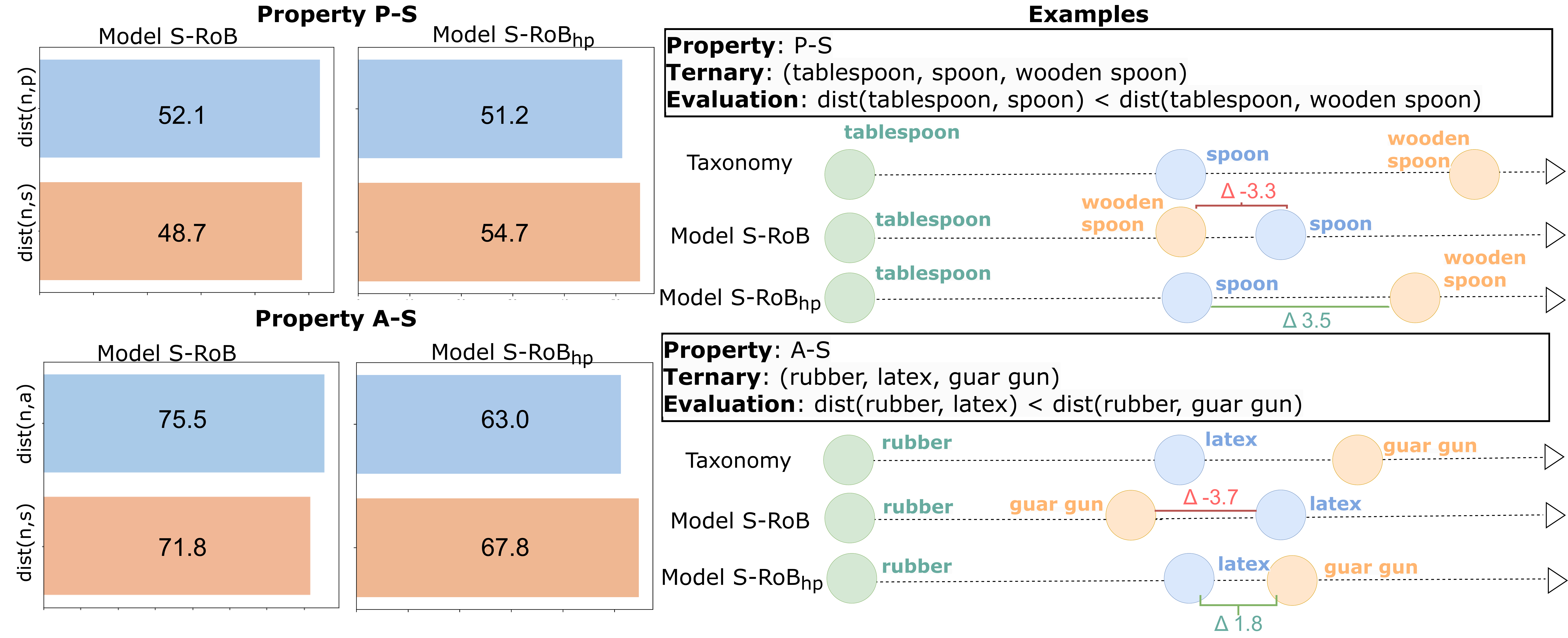}
    
    \label{fig:examples}
    \vspace{-0.8em}
\end{figure}

We further deepen our understanding of the impact of considering hierarchy on PLM's representations by comparing between \textbf{S-RoB} and \textbf{S-RoB$_{hp}$}. We particularly analyze the  failures, in terms of wrong predictions  of \textbf{S-RoB} using properties \textit{P-S} and \textit{A-S}. Figure \ref{fig:examples} (left) shows that both \textit{ancestor} and \textit{parent} relations initially exhibit greater distances, dist(n,p)$=52.1$ and dist(n,a)$=75.5$, for \textbf{S-RoB}, which subsequently decrease to $51.2$, $63.0$, respectively, after fine-tuning, for \textbf{S-RoB$_{hp}$}. Figure \ref{fig:examples} (right) shows hand-picked examples where we observe how the enhanced PLMs accurately reverse the trend of the distances between representations w.r.t. the evaluated properties.

To sum up, our experiments confirmed the feasibility of injecting hierarchy properties into PLMs, particularly for \textbf{BERT-L}, \textbf{S-RoB} and \textbf{S-MPNet} models, with overall performance higher than all the  evaluated vanilla PLMs. 

\addtolength{\textfloatsep}{-0.2in}

\vspace{-1em}
\subsection{Analyzing the transfer of hierarchy knowledge of PLMs to downstream tasks}
\vspace{-1em}
Finally, we answer \textbf{RQ3:} \textit{Can hierarchy-enhanced PLMs representations be transferred to downstream tasks?} We investigate in a sequential mode whether our hierarchy-enhanced PLMs using probes retain and leverage their knowledge when fine-tuned to perform a given downstream task.
%
In the following, we introduce each downstream task and report our results.
\vspace{-1.2em}
\subsubsection{Hypernym Discovery.}
This task evaluates an input term as a query and retrieves its suitable hypernyms from a target corpus. We used the SemEval-2018 Task 9 benchmark \cite{camacho-collados-etal-2018-semeval}.

Specifically, we consider the English subtask 1A which includes a vocabulary with all valid hypernyms, and a training/test set of annotated hypernym pairs. Due to the size of the corpus and queries, we employed a re-ranking approach involving two steps. 
Firstly, we considered two different models as first rankers to retrieve the top 1000 most relevant candidates: a semantic-retrieval approach with FT-wiki embeddings and cosine similarity, and BM25 with RM3 \cite{abdul2004umass}. Then, we re-ranked these filtered results using our hierarchy-enhanced and vanilla PLMs, which were fine-tuned with a binary classification layer using the CTP architecture  (for details, please refer to \cite{chen-etal-2021-CTP}).
We further explored an Oracle first ranker which extends the first ranker with missing golden candidates and acts as an upper bound of the performance of PLMs. We report MAP and P@5 ranking measures as proposed in \cite{camacho-collados-etal-2018-semeval}.

\vspace{-1.5em}

\subsubsection{Taxonomy Reconstruction.}
This task aims to construct a hierarchical taxonomy from a given set of words. We used the SemEval \textit{TexEval-II} \cite{bordea_2016} dataset\footnote{We used the English-language version of the taxonomies \textit{environment} and \textit{science}.} to compare with previous work \cite{chen-etal-2021-CTP,jain_2022}.
 The dataset consists of an evaluation set with edge-based accuracy as the metric. We followed CTP approach by training the vanilla and our hierarchy-enhanced PLMs on the Bansal dataset, omitting overlapping terms. We report standard Precision, Recall, and F1.

\vspace{-1.5em}
\subsubsection{Reading Comprehension.}
We use the RACE \cite{lai_2017} dataset, consisting of English exams for middle and high school Chinese students with up to four possible answers. The questions are split into Middle and High sets, where the High set is the most difficult. Reported values are given in terms of accuracy.

\begin{table}[tb] 
\scriptsize
    \caption{\small Results for vanilla and enhanced PLMs (left) and SOTA models (right). Hypernymy Discovery: report MAP and P@5. Taxonomy Reconstruction: report Precision (P), Recall(R), and F1. Reading Comprehension: report accuracy. Results are average of three runs and best values in \textbf{bold}.}
    \centering
    \resizebox{\textwidth}{!}{
\begin{tabular}{ccc|cc|cc|ccc|ccc}

\toprule
 &  \multicolumn{6}{c}{\textbf{Hypernym Discovery}}& \multicolumn{3}{c}{\textbf{Taxonomy}} & \multicolumn{3}{c}{\textbf{Reading}} \\ 
 &  \multicolumn{2}{c}{ \textbf{\footnotesize FT-wiki}} & \multicolumn{2}{c}{\textbf{\footnotesize BM25+RM3}} & \multicolumn{2}{c}{\textbf{Oracle}}&  \multicolumn{3}{c}{\textbf{Reconstruct.}} & \multicolumn{3}{c}{\textbf{Comprehension}}\\ \midrule
 \textbf{Model} &\textbf{MAP} &  \textbf{P@5} & \textbf{MAP} &  \textbf{P@5}& \textbf{MAP} &  \textbf{P@5}&  \textbf{P} & \textbf{R} & \textbf{F1} & \textbf{Middle} & \textbf{High} & \textbf{All}\\ \midrule
 BERT-L& 5.9&4.6&7.0&5.5&42.6&37.2&  15.5 & 48.0 & 22.9 & 75.0	&66.3	&68.8\\
RoB-L& 2.0&1.3&1.8&1.2&18.3&15.7 &  17.3 &  46.7 & 24.8 &  86.3&	80.0&	81.9\\
S-RoB& 10.4 &8.7 &11.6&9.8&45.5 &39.8&  17.2 & 33.6 & 22.6 & 57.9&	51.4	&53.3 \\
S-MPNet& 9.9&7.8&10.6&8.6&46.1&39.6 & 9.4 & 41.9 & 15.1 &  74.4	&67.8	&69.7\\
\midrule
BERT-L$_{hp}$& \textbf{11.3} & \textbf{9.7} &\textbf{12.4} & \textbf{11.0} & \textbf{53.6} & \textbf{48.5}   & 16.2 & 45.0 & 23.3 & 74.3	&63.4	&66.6\\
RoB-L$_{hp}$& 5.0 & 4.1 & 5.5 &4.5&32.2 & 27.0&  15.9\tiny &  \textbf{49.4} & 23.6 & \textbf{87.8}	&\textbf{81.4}	&\textbf{83.2}\\
S-RoB$_{hp}$& 9.2 & 7.8 & 10.5&9.1&45.6 & 39.9 &  18.3 & 32.8 & 23.3 &  57.3	&51.0	&52.8\\
S-MPNet$_{hp}$& 9.4 & 7.7 &10.5 &8.8&47.9 & 42.0& \textbf{41.4} & 27.1 & \textbf{32.3} & 46.2	&42.2	&43.4 \\
\bottomrule

\end{tabular} 
\quad
\begin{tabular}{ccccccc}
\toprule
\textbf{Hyper. Disc.}&  \multicolumn{3}{c}{\textbf{MAP}} & \multicolumn{3}{c}{ \textbf{P@5}} \\
\midrule
 BERT-L$_{hp}$    &\multicolumn{3}{c}{12.4} & \multicolumn{3}{c}{11.0} \\
  CRIM \cite{bernier-colborne-barriere-2018-crim}   & \multicolumn{3}{c}{$19.8$} &\multicolumn{3}{c}{$19.0$}\\
  RMM \cite{bai-etal-2021-hypernym} & \multicolumn{3}{c}{\textbf{27.1}} & \multicolumn{3}{c}{\textbf{23.4}}\\
  \midrule
\textbf{Taxonomy Reconst.}& \multicolumn{2}{c}{ \textbf{P}} &  \multicolumn{2}{c}{\textbf{R}} & \multicolumn{2}{c}{\textbf{F1}}\\
\midrule
 S-MPNet$_{hp}$   & \multicolumn{2}{c}{\textbf{41.4}} & \multicolumn{2}{c}{27.1} & \multicolumn{2}{c}{$32.3$}\\
  TaxoRL \cite{mao-etal-2018-end} & \multicolumn{2}{c}{$35.1$} & \multicolumn{2}{c}{\textbf{35.1}} & \multicolumn{2}{c}{\textbf{35.1}} \\
  CTP \cite{chen-etal-2021-CTP}  & \multicolumn{2}{c}{$26.3$} & \multicolumn{2}{c}{$25.9$} & \multicolumn{2}{c}{$26.1$}\\
  LMScorer \cite{jain_2022} & \multicolumn{2}{c}{$29.8$} & \multicolumn{2}{c}{$28.6$} & \multicolumn{2}{c}{$29.1$}   \\
  \midrule
\textbf{Reading Compr.}& \multicolumn{2}{c}{ \textbf{Middle}} &  \multicolumn{2}{c}{\textbf{High}} & \multicolumn{2}{c}{\textbf{All}}\\
\midrule
 RoB-L$_{hp}$   & \multicolumn{2}{c}{$87.8$} & \multicolumn{2}{c}{$81.4$} & \multicolumn{2}{c}{\textbf{$83.2$}}\\
 RoBERTa \cite{liu_2019} & \multicolumn{2}{c}{$86.5$} & \multicolumn{2}{c}{$81.8$} & \multicolumn{2}{c}{$82.8$}\\
 DeBERTA \cite{he2021debertav3} & \multicolumn{2}{c}{$90.5$} & \multicolumn{2}{c}{$86.8$} & \multicolumn{2}{c}{$87.5$}\\
 CoLISA\cite{colisa_2023}  & \multicolumn{2}{c}{\textbf{90.8}} & \multicolumn{2}{c}{\textbf{86.9}} & \multicolumn{2}{c}{\textbf{87.9}}   \\
  \bottomrule
\end{tabular}
}
    \label{tab:rq3}
\vspace{-0.5em}
\end{table}

We examine the transferability of the learned hierarchical representations to downstream tasks. Table \ref{tab:rq3} (left) shows results for all tasks. For Hypernym Discovery, our initial MAP scores were $2.2$, $3.4$, and $3.7$ for FT-wiki, BM25, and BM25+RM3, respectively. Additionally, re-rankers \textbf{BERT-L$_{hp}$} and \textbf{RoB-L$_{hp}$} improved $+5.4$ and $+5.1$ w.r.t. vanilla versions for MAP and P@5 based on BM25+RM3 initial ranking. However, we noticed a slight degradation in performance for \textbf{S-RoB$_{hp}$} and \textbf{S-MPNet$_{hp}$} (similarly for FT-wiki column). We also observed that all models improved their results for the Oracle results, with smaller margin ($+0.1$) for \textbf{S-RoB$_{hp}$} and \textbf{S-MPNet$_{hp}$}. These results motivate us to explore better first rankers. Moreover, considering the fact that this particular task heavily relies on hypernym relations (\textit{group} \textit{P-*}), we assume that the enhanced representations are easily transferable to this task. 
For Taxonomy Reconstruction, we report the average metrics from both taxonomies. Three models, \textbf{BERT-L$_{hp}$}, \textbf{S-RoB$_{hp}$}, and \textbf{S-MPNet$_{hp}$}, showed improvements in F1 scores w.r.t their vanilla version ($+0.4$, $+0.7$, and $+17.2$, respectively), while \textbf{ RoB-L$_{hp}$} degraded its performance by $-1.2$. The improvements were primarily driven by better precision ($0.7-32.0$), with a smaller penalization on the recall ($0.8-14.8$). Considering higher precision as an indication of better quality in the concept representations, we argue that the enhanced representations from \textit{groups} \textit{P-*, A-*}, and \textit{S-*} are transferable for this task. 
For Reading Comprehension, learned representations may not be fully transferable. Specifically, we observed improvements for \textbf{ RoB-L$_{hp}$}, but not for other models. This suggests that hierarchy knowledge is either forgotten \cite{doi:10.1073/pnas.1611835114} or penalizing for this task. Thus, an appropriate setup is called since a sequential fine-tuning might lead to the hierarchy representations drift for this task.

Table \ref{tab:rq3} (right) shows our results w.r.t SOTA models. For the tasks of Taxonomy Reconstruction and Reading Comprehension\footnote{All score recalculated for RoBERTa as suggested in the \href{https://github.com/facebookresearch/fairseq/issues/1565}{fairseq GitHub repository}.}, \textbf{S-MPNet$_{hp}$} and \textbf{ RoB-L$_{hp}$} achieve competitive performances. However, the main trend is that  specialized models, such as RRM, TaxoRL, and CoLISA, outperform fine-tuned enhanced models, which sheds light to the desirable room for improvement that a suitable transfer learning of hierarchy knowledge to downstream tasks, would achieve. Overall, our empirical findings suggest that the enhanced PLM representations are moderately transferable in a sequential mode of probe then fine-tune for hierarchy-aware tasks such as Hypernymy Discovery and Taxonomy Reconstruction, but detrimental for Reading Comprehension.

\vspace{-1.2em}
\section{Conclusion and Future Work}
\vspace{-1em}
In this paper, we proposed a task-agnostic methodology for probing the capability of PLMs to capture hierarchy.  We first identified hierarchy properties from taxonomies and then, constructed probes encoding these properties  to train hierarchy-enhanced PLMs.  Our experiments showed that cross-encoder models like BERT struggle to capture hierarchy and that hierarchy properties could be readily injected into PLMs regardless of specific tasks.  Evaluation of hierarchy-aware PLMs in downstream tasks reveals that a kind of catastrophic forgetting can occur leading to performance results under upper-bound performance of PLMs trained specifically on hierarchy-aware tasks.  

Future work will explore the use of transfer learning methods, such as adding hierarchy-aware auxiliary losses, that are able to learn robust features of a wide-range of hierarchy properties in a cross-task setting. This work is generalizable to other properties beyond hierarchy, leading to foster future research on the design of interpretable and effective PLMs for IR and NLP tasks. 

\newpage
%
%
%
%
\bibliographystyle{splncs04.bst}
\bibliography{iswc2023}

\begin{thebibliography}{10}
\providecommand{\url}[1]{\texttt{#1}}
\providecommand{\urlprefix}{URL }
\providecommand{\doi}[1]{https://doi.org/#1}

\bibitem{abdul2004umass}
Abdul-Jaleel, N., Allan, J., Croft, W.B., Diaz, F., Larkey, L., Li, X.,
  Smucker, M.D., Wade, C.: Umass at trec 2004: Novelty and hard. Computer
  Science Department Faculty Publication Series p.~189 (2004)

\bibitem{alsuhaibani2019joint}
Alsuhaibani, M., Maehara, T., Bollegala, D.: Joint learning of hierarchical
  word embeddings from a corpus and a taxonomy. In: Automated Knowledge Base
  Construction (AKBC) (2019)

\bibitem{bai-etal-2021-hypernym}
Bai, Y., Zhang, R., Kong, F., Chen, J., Mao, Y.: Hypernym discovery via a
  recurrent mapping model. In: Findings of the Association for Computational
  Linguistics: ACL-IJCNLP 2021. pp. 2912--2921. Online (Aug 2021)

\bibitem{bansal_2014}
Bansal, M., Burkett, D., {de Melo}, G., Klein, D.: Structured {{Learning}} for
  {{Taxonomy Induction}} with {{Belief Propagation}}. In: ACL 2014. pp.
  1041--1051 (2014)

\bibitem{bernier-colborne-barriere-2018-crim}
Bernier-Colborne, G., Barri{\`e}re, C.: {CRIM} at {S}em{E}val-2018 task 9: A
  hybrid approach to hypernym discovery. In: Proceedings of the 12th
  International Workshop on Semantic Evaluation. pp. 725--731. New Orleans,
  Louisiana (Jun 2018)

\bibitem{bojanowski_2017}
Bojanowski, P., Grave, E., Joulin, A., Mikolov, T.: Enriching {{Word Vectors}}
  with {{Subword Information}}. TACL 2017  \textbf{5},  135--146 (2017)

\bibitem{bordea_2015}
Bordea, G., Buitelaar, P., Faralli, S., Navigli, R.: {{SemEval-2015 Task}} 17:
  {{Taxonomy Extraction Evaluation}} ({{TExEval}}). In: SemEval-2015. pp.
  902--910 (2015)

\bibitem{bordea_2016}
Bordea, G., Lefever, E., Buitelaar, P.: {{SemEval-2016 Task}} 13: {{Taxonomy
  Extraction Evaluation}} ({{TExEval-2}}). In: SemEval-2016. pp. 1081--1091
  (2016)

\bibitem{bordes_2013}
Bordes, A., Usunier, N., Garcia-Duran, A., Weston, J., Yakhnenko, O.:
  Translating embeddings for modeling multi-relational data. In: Advances in
  Neural Information Processing Systems (NIPS). vol.~26 (2013)

\bibitem{budanitsky_2006}
Budanitsky, A., Hirst, G.: Evaluating {{WordNet-based Measures}} of {{Lexical
  Semantic Relatedness}}. Computational Linguistics  \textbf{32}(1),  13--47
  (2006)

\bibitem{camacho-collados_2017}
{Camacho-Collados}, J.: Why we have switched from building full-fledged
  taxonomies to simply detecting hypernymy relations (Mar 2017)

\bibitem{camacho-collados-etal-2018-semeval}
Camacho-Collados, J., Delli~Bovi, C., Espinosa-Anke, L., Oramas, S., Pasini,
  T., Santus, E., Shwartz, V., Navigli, R., Saggion, H.: {S}em{E}val-2018 task
  9: Hypernym discovery. In: Proceedings of the 12th International Workshop on
  Semantic Evaluation. pp. 712--724. Association for Computational Linguistics,
  New Orleans, Louisiana (Jun 2018)

\bibitem{camara2020diagnosing}
C{\^a}mara, A., Hauff, C.: Diagnosing bert with retrieval heuristics. In:
  Advances in Information Retrieval: 42nd European Conference on IR Research,
  ECIR 2020, Lisbon, Portugal, April 14--17, 2020, Proceedings, Part I 42. pp.
  605--618. Springer (2020)

\bibitem{chen2020probing}
Chen, B., Fu, Y., Xu, G., Xie, P., Tan, C., Chen, M., Jing, L.: Probing bert in
  hyperbolic spaces. In: International Conference on Learning Representations
  (2020)

\bibitem{chen-etal-2021-CTP}
Chen, C., Lin, K., Klein, D.: Constructing taxonomies from pretrained language
  models. In: Proceedings of the 2021 Conference of the North American Chapter
  of the Association for Computational Linguistics: Human Language Technologies
  (NAACL). pp. 4687--4700. Online (Jun 2021)

\bibitem{clark_2020}
Clark, P., Richardson, K., Tafjord, O.: Transformers as {{Soft Reasoners}} over
  {{Language}}. In: IJCAI 2020. vol.~4, pp. 3882--3890 (2020)

\bibitem{devlin_2019}
Devlin, J., Chang, M.W., Lee, K., Toutanova, K.: {{BERT}}: {{Pre-training}} of
  {{Deep Bidirectional Transformers}} for {{Language Understanding}}. In: NAACL
  2019. pp. 4171--4186 (2019)

\bibitem{colisa_2023}
Dong, M., Zou, B., Li, Y., Hong, Y.: Colisa: Inner interaction via contrastive
  learning for multi-choice reading comprehension. In: Advances in Information
  Retrieval: 45th European Conference on Information Retrieval, ECIR 2023,
  Dublin, Ireland, April 2–6, 2023, Proceedings, Part I. p. 264–278.
  Springer-Verlag, Berlin, Heidelberg (2023).
  \doi{10.1007/978-3-031-28244-7_17}

\bibitem{dong_2018a}
Dong, X., Shen, J.: Triplet {{Loss}} in {{Siamese Network}} for {{Object
  Tracking}}. In: ECCV 2018. pp. 459--474 (2018)

\bibitem{du2022hakg}
Du, Y., Zhu, X., Chen, L., Zheng, B., Gao, Y.: Hakg: Hierarchy-aware knowledge
  gated network for recommendation. In: Proceedings of the 45th International
  ACM SIGIR Conference on Research and Development in Information Retrieval.
  pp. 1390--1400 (2022)

\bibitem{fu_2014}
Fu, R., Guo, J., Qin, B., Che, W., Wang, H., Liu, T.: Learning {{Semantic
  Hierarchies}} via {{Word Embeddings}}. In: ACL 2014. pp. 1199--1209 (2014)

\bibitem{gupta_2016}
Gupta, A., Piccinno, F., Kozhevnikov, M., Pa{\c s}ca, M., Pighin, D.:
  Revisiting {{Taxonomy Induction}} over {{Wikipedia}}. In: COLING 2016. pp.
  2300--2309 (2016)

\bibitem{he2021debertav3}
He, P., Gao, J., Chen, W.: Debertav3: Improving deberta using electra-style
  pre-training with gradient-disentangled embedding sharing. arXiv preprint
  arXiv:2111.09543  (2021)

\bibitem{hovy-etal-2009-toward}
Hovy, E., Kozareva, Z., Riloff, E.: Toward completeness in concept extraction
  and classification. In: Proceedings of the 2009 Conference on Empirical
  Methods in Natural Language Processing (EMNLP). pp. 948--957 (Aug 2009)

\bibitem{hu_2015}
Hu, Z., Huang, P., Deng, Y., Gao, Y., Xing, E.: Entity {{Hierarchy Embedding}}.
  In: IJCNLP 2015. pp. 1292--1300 (2015)

\bibitem{jain_2022}
Jain, D., Anke, L.E.: Distilling {{Hypernymy Relations}} from {{Language
  Models}}: {{On}} the {{Effectiveness}} of {{Zero-Shot Taxonomy Induction}}.
  arXiv:2202.04876 [cs]  (2022)

\bibitem{jain-starsem-22}
Jain, D., Espinosa~Anke, L.: Distilling hypernymy relations from language
  models: On the effectiveness of zero-shot taxonomy induction. In: Proceedings
  of the 11th Joint Conference on Lexical and Computational Semantics. pp.
  151--156. Seattle, Washington (Jul 2022)

\bibitem{ji-etal-2015-knowledge}
Ji, G., He, S., Xu, L., Liu, K., Zhao, J.: Knowledge graph embedding via
  dynamic mapping matrix. In: Proceedings of the 53rd Annual Meeting of the
  Association for Computational Linguistics and the 7th International Joint
  Conference on Natural Language Processing (NAACL). pp. 687--696. Beijing,
  China (Jul 2015)

\bibitem{doi:10.1073/pnas.1611835114}
Kirkpatrick, J., Pascanu, R., Rabinowitz, N., Veness, J., Desjardins, G., Rusu,
  A.A., Milan, K., Quan, J., Ramalho, T., Grabska-Barwinska, A., Hassabis, D.,
  Clopath, C., Kumaran, D., Hadsell, R.: Overcoming catastrophic forgetting in
  neural networks. Proceedings of the National Academy of Sciences
  \textbf{114}(13),  3521--3526 (2017). \doi{10.1073/pnas.1611835114},
  \url{https://www.pnas.org/doi/abs/10.1073/pnas.1611835114}

\bibitem{lai_2017}
Lai, G., Xie, Q., Liu, H., Yang, Y., Hovy, E.: {{RACE}}: {{Large-scale ReAding
  Comprehension Dataset From Examinations}}. In: EMNLP 2017. pp. 785--794
  (2017)

\bibitem{liu2023seeking}
Liu, C., Cohn, T., Frermann, L.: {Seeking Clozure: Robust Hypernym extraction
  from BERT with Anchored Prompts}. In: Proceedings of the The 12th Joint
  Conference on Lexical and Computational Semantics (* SEM 2023). pp. 193--206
  (2023)

\bibitem{liu2019kbert}
Liu, W., Zhou, P., Zhao, Z., Wang, Z., Ju, Q., Deng, H., Wang, P.: K-bert:
  Enabling language representation with knowledge graph. In: Proceedings of the
  AAAI Conference on Artificial Intelligence. vol.~34, pp. 2901--2908 (2020)

\bibitem{liu_2019}
Liu, Y., Ott, M., Goyal, N., Du, J., Joshi, M., Chen, D., Levy, O., Lewis, M.,
  Zettlemoyer, L., Stoyanov, V.: {{RoBERTa}}: {{A Robustly Optimized BERT
  Pretraining Approach}}. arXiv:1907.11692 [cs]  (2019)

\bibitem{mao-etal-2018-end}
Mao, Y., Ren, X., Shen, J., Gu, X., Han, J.: End-to-end reinforcement learning
  for automatic taxonomy induction. In: Proceedings of the 56th Annual Meeting
  of the Association for Computational Linguistics (ACL). pp. 2462--2472.
  Melbourne, Australia (Jul 2018)

\bibitem{nickel_2017}
Nickel, M., Kiela, D.: Poincar\'e {{Embeddings}} for {{Learning Hierarchical
  Representations}}. In: NIPS 2017. vol.~30 (2017)

\bibitem{pimentel-etal-2020-information}
Pimentel, T., Valvoda, J., Hall~Maudslay, R., Zmigrod, R., Williams, A.,
  Cotterell, R.: Information-theoretic probing for linguistic structure. In:
  Proceedings of the 58th Annual Meeting of the Association for Computational
  Linguistics (ACL) (2020)

\bibitem{Qiu-2020}
Qiu, X., Sun, T., Xu, Y., Shao, Y., Dai, N., Huang, X.: Pre-trained models for
  natural language processing: {A} survey. CoRR  \textbf{abs/2003.08271}
  (2020), \url{https://arxiv.org/abs/2003.08271}

\bibitem{reimers_2019}
Reimers, N., Gurevych, I.: Sentence-{{BERT}}: {{Sentence Embeddings}} using
  {{Siamese BERT-Networks}}. In: EMNLP-IJCNLP (2019)

\bibitem{resnik_1995}
Resnik, P.: Using information content to evaluate semantic similarity in a
  taxonomy. In: IJCAI 1995. pp. 448--453 (1995)

\bibitem{richardson_2020}
Richardson, K., Sabharwal, A.: What {{Does My QA Model Know}}? {{Devising
  Controlled Probes}} using {{Expert}}. TACL 2020  \textbf{8}(0),  572--588
  (2020)

\bibitem{rossi_2021}
Rossi, A., Barbosa, D., Firmani, D., Matinata, A., Merialdo, P.: Knowledge
  graph embedding for link prediction: A comparative analysis. ACM Trans.
  Knowl. Discov. Data  \textbf{15}(2) (jan 2021)

\bibitem{sieg2004}
Sieg, A., Mobasher, B., Lytinen, S., Burke, R.: Using concept hierarchies to
  enhance user queries in web-based information retrieval. Artificial
  Intelligence and Applications (AIA)  (2004)

\bibitem{song2022hyperbolic}
Song, M., Feng, Y., Jing, L.: Hyperbolic relevance matching for neural
  keyphrase extraction. In: Proceedings of the 2022 Conference of the North
  American Chapter of the Association for Computational Linguistics: Human
  Language Technologies. pp. 5710--5720 (2022)

\bibitem{Sun_Wang_Li_Feng_Tian_Wu_Wang_2020}
Sun, Y., Wang, S., Li, Y., Feng, S., Tian, H., Wu, H., Wang, H.: Ernie 2.0: A
  continual pre-training framework for language understanding. Proceedings of
  the AAAI Conference on Artificial Intelligence  \textbf{34}(05),  8968--8975
  (Apr 2020). \doi{10.1609/aaai.v34i05.6428},
  \url{https://ojs.aaai.org/index.php/AAAI/article/view/6428}

\bibitem{talmor_2020}
Talmor, A., Elazar, Y., Goldberg, Y., Berant, J.: {{oLMpics-On What Language
  Model Pre-training Captures}}. TACL 2020  \textbf{8},  743--758 (2020)

\bibitem{vendrov_2016a}
Vendrov, I., Kiros, R., Fidler, S., Urtasun, R.: Order-embeddings of images and
  language. In: 4th International Conference on Learning Representations,
  {ICLR} 2016, San Juan, Puerto Rico,2016, Conference Track Proceedings (2016)

\bibitem{vulic_2021}
Vuli{\'c}, I., Ponti, E.M., Korhonen, A., Glava{\v s}, G.: {{LexFit}}:
  {{Lexical Fine-Tuning}} of {{Pretrained Language Models}}. In: ACL 2021. pp.
  5269--5283 (2021)

\bibitem{vulic_2020}
Vuli{\'c}, I., Ponti, E.M., Litschko, R., Glava{\v s}, G., Korhonen, A.:
  Probing {{Pretrained Language Models}} for {{Lexical Semantics}}. In: EMNLP
  2020. pp. 7222--7240 (2020)

\bibitem{wang-etal-2021-tsdae-using}
Wang, K., Reimers, N., Gurevych, I.: {TSDAE}: Using transformer-based
  sequential denoising auto-encoderfor unsupervised sentence embedding
  learning. In: Findings of the Association for Computational Linguistics:
  EMNLP 2021. pp. 671--688. Association for Computational Linguistics, Punta
  Cana, Dominican Republic (Nov 2021)

\bibitem{yang2022survey}
Yang, J., Xiao, G., Shen, Y., Jiang, W., Hu, X., Zhang, Y., Peng, J.: A survey
  of knowledge enhanced pre-trained models (2022)

\bibitem{zhang_2019}
Zhang, Z., Han, X., Liu, Z., Jiang, X., Sun, M., Liu, Q.: {{ERNIE}}: {{Enhanced
  Language Representation}} with {{Informative Entities}}. In: ACL 2019. pp.
  1441--1451 (2019)

\bibitem{zhong_2002}
Zhong, J., Zhu, H., Li, J., Yu, Y.: Conceptual graph matching for semantic
  search. In: In {{ICCS}}. pp. 92--106 (2002)

\end{thebibliography}

\end{document}